\documentclass[conference]{IEEEtran}
\usepackage{times}

\usepackage{amsmath,amsfonts,bm}

\def\Figref#1{Figure~\ref{#1}}

\def\Tabref#1{Table~\ref{#1}}

\def\Secref#1{Section~\ref{#1}}

\def\eqref#1{equation~\ref{#1}}

\def\1{\bm{1}}

\DeclareMathAlphabet{\mathsfit}{\encodingdefault}{\sfdefault}{m}{sl}
\SetMathAlphabet{\mathsfit}{bold}{\encodingdefault}{\sfdefault}{bx}{n}

\usepackage[numbers]{natbib}
\usepackage{multicol}

\usepackage{url}
\usepackage[pdftex]{graphicx}
\usepackage{multirow}
\usepackage{enumitem}
\usepackage[font={small}]{caption}
\usepackage{natbib}

\begin{document}

\title{Swoosh! Rattle! Thump! - Actions that Sound}


\author{\authorblockN{Dhiraj Gandhi}
\authorblockA{CMU\\
dgandhi@cs.cmu.edu
}
\and
\authorblockN{Abhinav Gupta}
\authorblockA{CMU\\
abhinavg@cs.cmu.edu}
\and
\authorblockN{Lerrel Pinto \hspace{0.16in}}
\authorblockA{CMU / NYU \hspace{0.12in} \\
lerrel@cs.nyu.edu \hspace{0.14in}} 
}

\makeatletter
\let\@oldmaketitle\@maketitle%
\renewcommand{\@maketitle}{\@oldmaketitle%
    \centering
    \includegraphics[width=\linewidth]{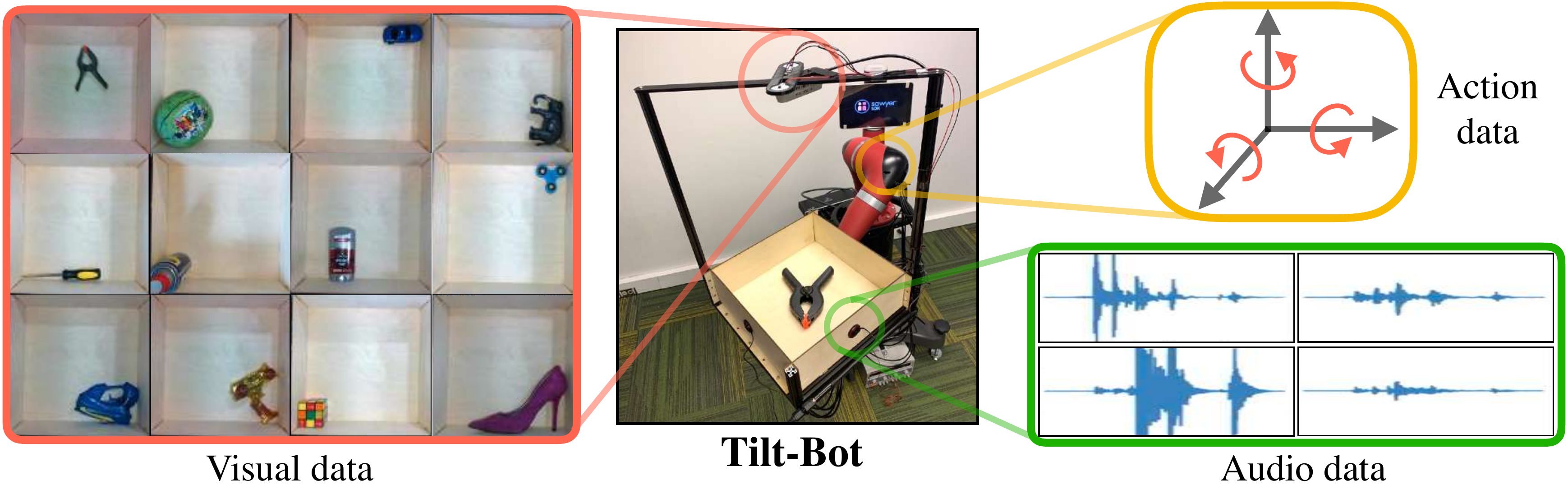}
    \vspace{0.01in}
    \captionof{figure}{Using Tilt-Bot, we collect 15,000 interactions on 60 different objects by tilting them in a tray. When sufficiently tilted, the object slides across the tray and hits the walls of the tray. This generates sound, which is captured by four contact microphones mounted on each side of the tray. An overhead camera records visual (RGB+Depth) information, while the robotic arm applies the tilting actions through end-effector rotations.}
    \label{fig:intro}
}
\makeatother
\maketitle
\begin{abstract}
Truly intelligent agents need to capture the interplay of all their senses to build a rich physical understanding of their world. In robotics, we have seen tremendous progress in using visual and tactile perception; however, we have often ignored a key sense: sound. This is primarily due to the lack of data that captures the interplay of action and sound. In this work, we perform the first large-scale study of the interactions between sound and robotic action. To do this, we create the largest available sound-action-vision dataset with 15,000 interactions on 60 objects using our robotic platform Tilt-Bot. By tilting objects and allowing them to crash into the walls of a robotic tray, we collect rich four-channel audio information. Using this data, we explore the synergies between sound and action and present three key insights. First, sound is indicative of fine-grained object class information, e.g., sound can differentiate a metal screwdriver from a metal wrench. Second, sound also contains information about the causal effects of an action, i.e. given the sound produced, we can predict what action was applied to the object. Finally, object representations derived from audio embeddings are indicative of implicit physical properties. We demonstrate that on previously unseen objects, audio embeddings generated through interactions can predict forward models 24\% better than passive visual embeddings. 
Project videos and data are at 
https://dhiraj100892.github.io/swoosh/
\end{abstract}

\IEEEpeerreviewmaketitle

\section{Introduction}
Imagine the opening of a champagne bottle! Most vivid imaginations not only capture the celebratory visuals but also the distinctive `pop' sound created by the act. Our world is rich and feeds all of our five senses -- vision, touch, smell, sound, and taste. Of these, the senses of vision, touch, and sound play a critical role in our rich physical understanding of objects and actions. A truly intelligent agent would need to capture the interplay of all the three senses to build a physical understanding of the world. In robotics, where the goal is to perform a physical task, the vision has always played a central role. Vision is used to infer the geometric shape~\cite{shapesKarTCM15}, track objects~\cite{xiang2015learning}, infer object categories~\cite{krizhevsky2012imagenet} and even direct control~\cite{levine2016end}. In recent years, the sense of touch has also received increasing attention for recognition~\cite{schneider2009object} and feedback control~\cite{DBLP:journals/corr/abs-1805-04201}. But what about sound? From the squeak of a door to the rustle of a dried leaf, sound captures rich object information that is often imperceptible through visual or force data. Microphones (sound sensors) are also inexpensive and robust, yet we haven't  seen sound data transform robot learning.
There hardly exist any systems, algorithms, or datasets that exploit sound as a vehicle to build physical understanding. Why is that? Why does sound appear to be a second-class citizen among perceptual faculties?


The key reason lies at the heart of sound generation. Sound generated through an interaction, say a robot striking an object, depends on the impact of the strike, the structure of the object, and even the location of the microphone. This intricate interplay that generates rich data also makes it difficult to extract information that is useful for robotics. Although recent work by \citet{pmlr-v87-clarke18a} has used sound to determine the amount of granular material in a container, we believe there lies much more information in the sound of interactions. But what sort of information can be extracted from this sound?

In this paper, we explore the synergy between sound and action to gain insight into what sound can be used for. To begin this exploration we will first need a large and diverse dataset that contains both sound and action data. However, most existing sound datasets do not contain information about the action, while most action datasets do not contain information about sound. 
To solve this, we create the largest sound-action-vision dataset available with 15,000 interactions on over 60 objects with our Tilt-Bot robot ~\Figref{fig:intro}. Each object is placed in a tray mounted on a robot arm that is tilted with a random action until the object hits the walls of the tray and makes a sound. This setup allows us to robustly collect sound and action data over a diverse set of objects. But how is this data useful? Through Tilt-Bot's data, we present three key insights about the role of sound in action.

The first insight is that sound is indicative of fine-grained object information. This implies that just from the sound an object makes, a learned model can identify the object with 79.2\% accuracy from a set of diverse 60 objects, which includes 30 YCB objects~\cite{calli2015benchmarking}. Our second insight is that sound is indicative of action. This implies that just from hearing the sound of an object, a learned model can predict what action was applied to the object. On a set of 30 previously unseen objects, we achieve a 0.027 mean squared error which is 42\% better than learning from only visual inputs. 
Our final insight is that sound is indicative of the physical properties of an object. This implies that just from hearing the sound an object makes, a learned model can infer the implicit physical properties of the object. To test this implicit physics, we show that a learned audio-conditioned forward model achieves a L1 error of 0.193 on previously unseen objects, which is 24\% lesser than forward models trained using visual information. This further indicates that audio embeddings, generated from a previous interaction, can capture information about the physics of an object significantly better than visual embeddings. One could envision using these features to learn policies that first interact to create sound and then use the inferred audio embeddings to perform actions~\cite{zhou2019environment}.


In summary, we present three key contributions in this paper: (a) we create the largest sound-action-vision robotics dataset; (b) we demonstrate that we can perform fine grained object recognition using only sound; and (c) we show that sound is indicative of action, both for post-interaction prediction, and pre-interaction forward modeling. Tilt-Bot's sound-action-vision data, along with audio embeddings will be publicly released.

\section{Related work}
\begin{figure*}[t!]
\centering
\includegraphics[width=\textwidth]{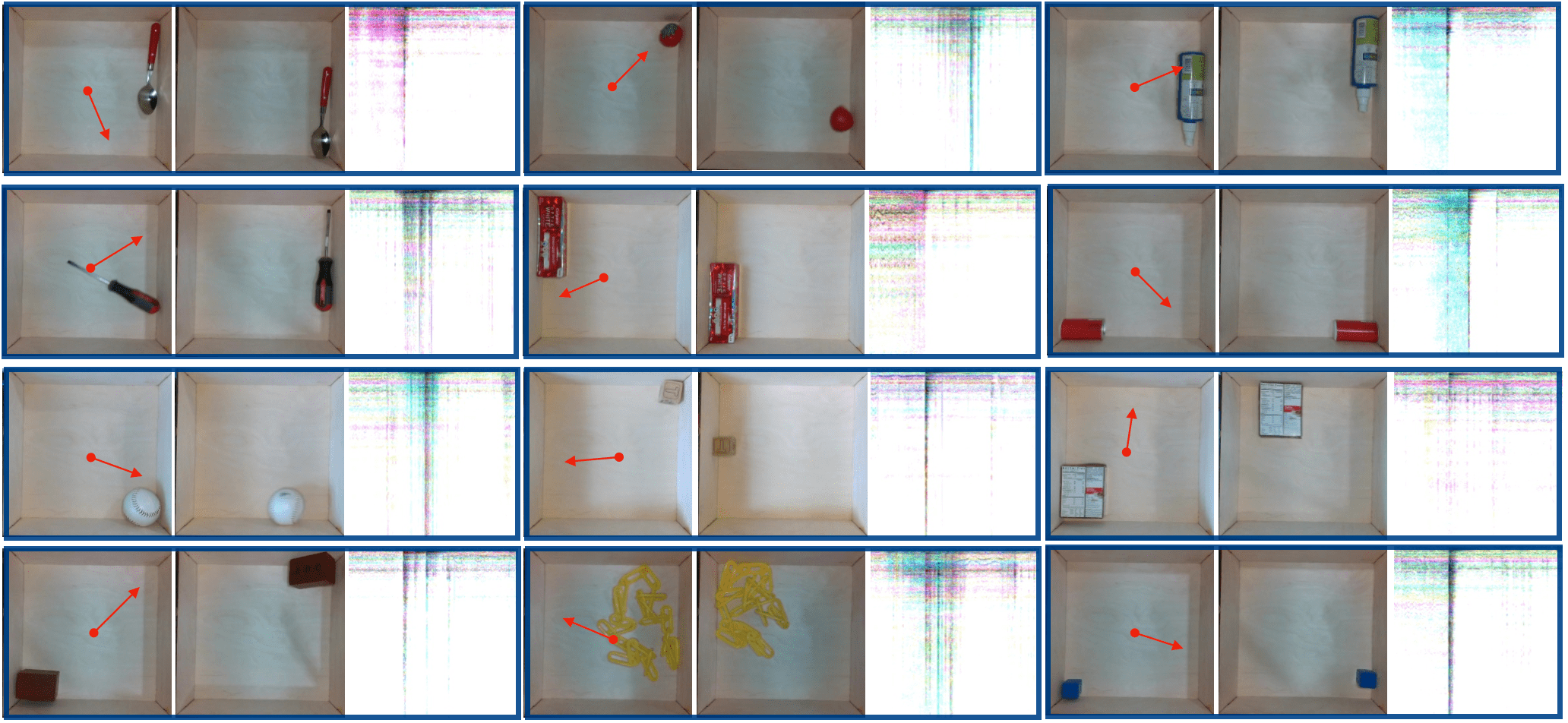}
\caption{Here are 12 interactions collected using Tilt-Bot. Each interaction is visualized as three images. The left image shows the visual observation of the object before the action is applied along with the applied action. The middle image shows the visual observation after the interaction, while the right image shows the STFT representation of audio generated by 3 of the 4 microphones.}
\label{fig:data}
\end{figure*}

\subsection{Multi-modal Learning with Sound}

Recently, there has been a growing interest to use sound in conjunction with vision, either for generating sound for mute videos, or to localize the part of the image that produces sound, or to learn better visual and audio features. 
For instance, \citet{andrew_sound} collected hundreds  of  videos  of  people  hitting, scratching, and prodding objects with a drumstick. This data was then used to train a recurrent neural network which synthesizes sound from silent videos. 
However, relying on humans for data collection is a considerable bottleneck to scale up the approach.  Similarly, \citet{zhang2017shape} build a simulator to generate sounds for different objects and then used it to learn the physical properties of objects. However, due to the  domain gap between simulation and real, widespread application of these techniques has been limited. \citet{carl_soundnet} uses the natural synchronization between vision and sound to learn an acoustic representation using two-million unlabelled videos. Similarly, \citet{andrew_look_listen_learn} looks at raw unconstrained videos to learn visual and audio representations that perform on par with state-of-the-art self-supervised approaches. In a similar spirit, we also learn audio representations, albeit through action, to be used for downstream tasks.  \citet{andrew_objects_that_sound, DBLP:journals/corr/abs-1803-03849} further explores the audio-visual correspondence in videos to localize the object that sounds in an image, given the audio signal. \citet{Zhao_2018_ECCV} has taken this idea one step further. Given the audio signal, they separate it into a set of components that represents the sound from each pixel. In contrast to these works, we look at obtaining a richer representation of sound by studying its interactions with action.


\subsection{Learning forward models}
Standard  model-based  methods, based on the  estimated physical  properties and  known laws of physics, try to calculate a sequence of control actions to achieve the goal~\cite{Khatib1987AUA,murray2017mathematical,dogar2012planning}. Although this has been widely used for object manipulation tasks in robotics~\cite{cosgun2011push}, manipulating an unknown object is still a daunting task for such methods. This is mainly due to the difficulties  in estimating and modeling the novel physical world~\cite{yu2016more}. Given the challenges in predicting the physical properties of a novel environment, several works~\cite{deisenroth2011pilco, gal2016improving, amos2018learning, henaff2017model} try to learn dynamic models based on objects' interactions with the environment. However, when we need to use these learned models on previously unseen objects, these models also fail to generalize. This is because they often do not contain object-specific information. One way to get object-specific information is to use raw visual observations instead of object state~\cite{agarwal2016, hafner2018learning, finn2017deep}. In these methods, given the observation of a scene and action taken by the agent, a visual forward model predicts the future visual observation. These forward models can then be used to plan robotic motions. In our work, we show that instead of using visual information, audio embeddings generated from a previous interaction can be used to improve these forward models. 

\subsection{Multi-modal Datasets}
Alongside algorithmic developments, large scale datesets have enabled the application of machine learning to solve numerous robotic tasks. Several works like \citet{pinto2016supersizing, levine2016learning,agarwal2016,DBLP:journals/corr/GandhiPG17} collect large scale visual robotic data for learning manipulation and navigation skills. Apart from visual data, some works~\cite{DBLP:journals/corr/abs-1805-04201, pinto2016curious} have also looked at collecting large-scale tactile data. This tactile or force data can then be used to recover object properties like softness or roughness. Although these datasets contain visual information and action data, they ignore a key sensory modality: sound. 

Understanding what information can be obtained from sound requires a large-scale sound dataset. 
Early work~\cite{andrew_sound} collected sound data by recording people interacting with objects. 
Although this dataset contains large amounts of sound data, it does not contain information about the action. In our work, we show that action information not only helps regularize object classification but also helps in understanding the implicit physics of objects. Prior to our work, \citet{pmlr-v87-clarke18a} has shown that sound information is indeed helpful for state-estimation tasks like measuring the amount of granular material in a container. Here, they exploit the mechanical vibrations of granular material and the structure around it for accurate estimation. In our work, instead of a single type of object, we collect audio data across 60 different objects. This allows us to learn generalizable audio features that transfer to previously unseen objects on a variety of tasks like action regression and forward-model learning.

\section{The Tilt-Bot Sound Dataset}

To study the relationship between sound and actions, we first need to create a dataset with sound and action. In this section, we describe our data collection setup and other design decisions.

\noindent \textbf{The Tilt-Bot Setup:} A framework to collect large-scale data needs three key abilities: (a) to precisely control the actions; (b) to be able to interact with a diverse set of objects; (c) to record rich and diverse sound; and (d) requires little to no manual resets. To do this, we present Tilt-Bot ~(\Figref{fig:intro}). Tilt-Bot is a robotic tray mounted on a Sawyer robot's end-effector. This allows us to precisely control the movements of the tray by applying rotational and translational actions on objects inside it. The tray has dimensions of $30\times30$ cm and a payload of 1 Kg allowing us to place a large range of common day objects in it. To collect audio information, four contact microphones ~\cite{microphone} are attached on the four sides of the tray. This allows for the creation of rich audio information from the interactions of objects with each other and the tray. To collect visual information, an Intel Realsense Camera ~\cite{camera} is mounted on the top of the tray to give RGB and Depth information of the object in the tray. Our current setup allows us to collect four-channel audio at 44,100Hz, RGB and Depth at 6Hz, and tray state information (rotation and translation) at 100Hz. Rotational and translational action commands can be sent at 100Hz. 
\begin{figure*}[t!]
\centering
\includegraphics[width=\linewidth]{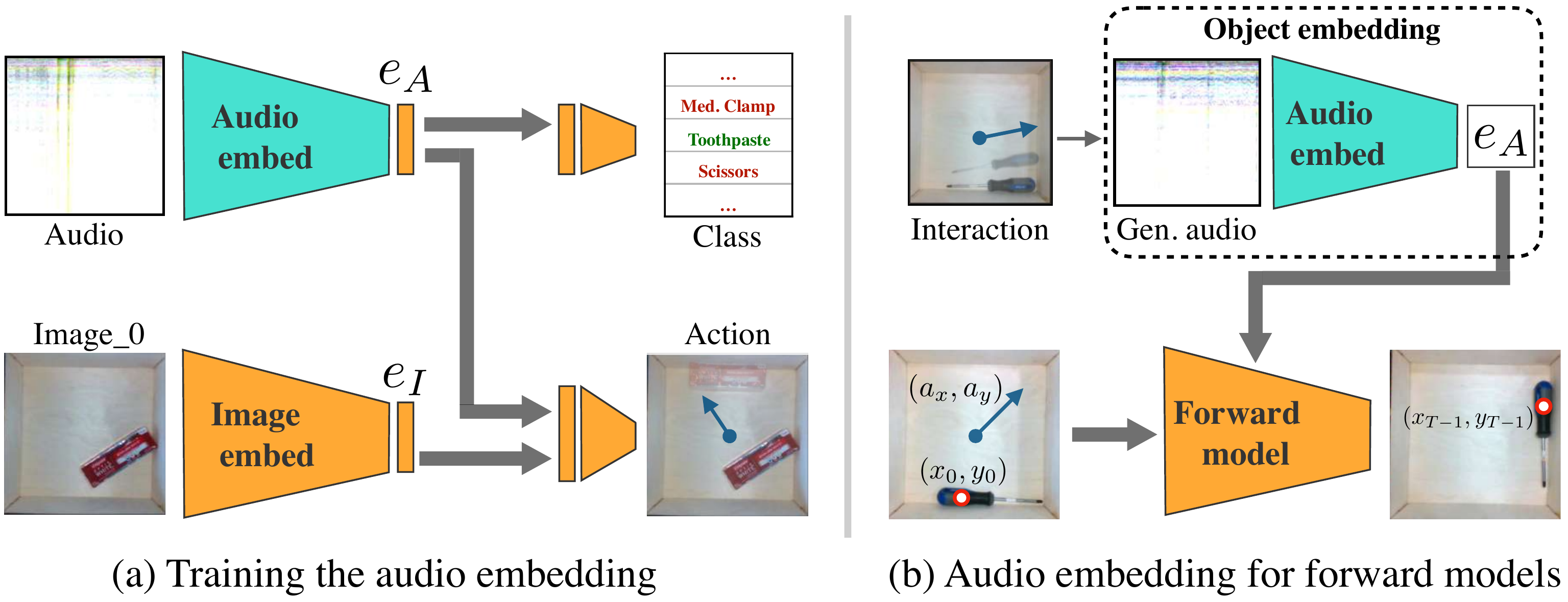}
\caption{To train audio embeddings (a), we perform multi-task learning for instance recognition (top) and action regression (bottom). Once the embedding network is trained, we can use the extracted audio embeddings as object features for downstream tasks like forward model learning (b).}
\label{fig:nets}
\end{figure*}
\begin{figure}[h]
\centering
\begin{minipage}{.5\linewidth}
  \begin{center}
        \includegraphics[width=0.95\linewidth]{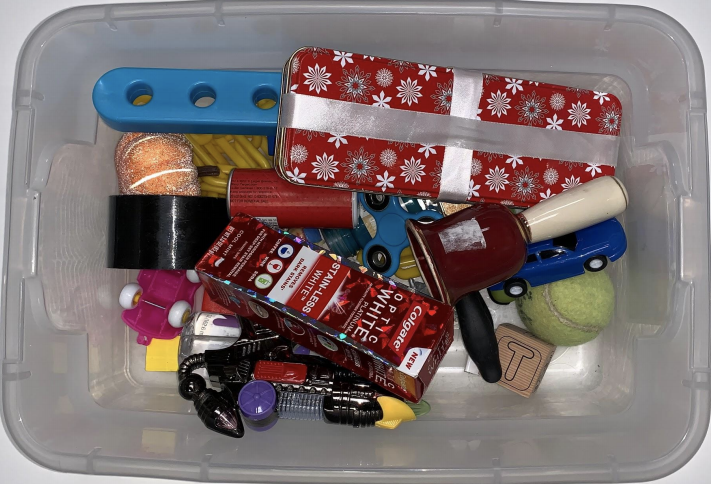}
        \caption*{Training (set A) Objects}
  \end{center}
\end{minipage}%
\begin{minipage}{.5\linewidth}
  \begin{center}
        \includegraphics[width=0.95\linewidth]{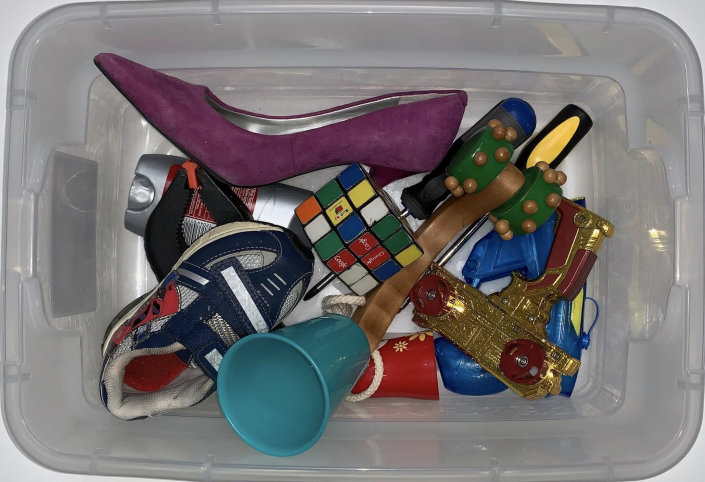}
        \caption*{Testing (set B) Objects}
  \end{center}
\end{minipage}
\caption{Images of some of the object used for collecting data using Tilt-Bot. }
\label{fig:objects_image}
\end{figure}

\noindent \textbf{Data Collection Procedure:} Our dataset consists of sound-action-vision data on 60 objects; 30 of which belong to the YCB object dataset ~\cite{calli2015benchmarking}, and 30 are common household objects. ~\Figref{fig:objects_image} represents some of the objects used for collecting data using Tilt-Bot. For each object, data is collected by first placing it in the center of the tray. Then, Tilt-Bot applies randomly generated rotational actions to the object for 1 hour. We do not apply translational action since we notice the minimal motion of the object with it. The rotational actions cause the tray to tilt and make the object slide and hit the walls of the tray. The sound from the four microphones, along with the visual data are continually recorded. Furthermore, using a simple background subtraction technique~\cite{zivkovic2004improved}, we can track the location of the object as it collides with the walls of the tray. For every contact made with the tray's wall, which is detected by peaks in the audio stream, we segment a four-second interaction centered around this contact. This amounts to around 15000 interactions over 60 robotic hours of data collection. Each of these interactions contains the sound, the RGB+Depth, and the tracking location of the object during the interaction. Examples of the data can be seen in ~\Figref{fig:data}. All of our data and pre-processing will be open-sourced.

\section{Learning with Audio}

To understand and study the synergies between sound and action, we focus on three broad categories of learning tasks: (1) fine-grained classification (or instance recognition)~\cite{krizhevsky2012imagenet, He_2017_ICCV}, (2) inverse-model learning (or action regression)~\cite{pinto2016supersizing, agarwal2016}, and (3) downstream forward-model learning ~\cite{ebert2018visual, deisenroth2011pilco}. In this section, we will describe our experiments along with insights to better understand the role of sound with action in the context of learning and robotics.

\subsection{Processing audio data}
Before using audio data for learning, we first need to convert it into a canonical form. Since we will use audio in conjunction with images for several experiments, we build on the representation proposed by \citet{Zhao_2018_ECCV}. Here the key idea is to convert the high dimensional raw audio (705600 for a 4 second audio recorded at 44.1KHz for 4 audio channels) to a smaller dimensional image. This is first done by subsampling each audio channel from 44.1KHz to 11KHz. Then, a Short-time Fourier transform (STFT)~\cite{daubechies1990wavelet} with a FFT window size of 510 and hop length of 128 is applied on the subsampled and clipped audio data. For each channel this results in a $64\times64$ representation. Stacking the 4 channel audio, we get a $64\times64\times4$ representation. We further apply a log transformation and clip the representation to between $\lbrack -5, 5\rbrack$. This representation allows us to treat audio as an image and now effectively run 2D convolutions on audio data, which can capture the temporal correlations from a single audio channel along with the correlations between multiple audio channels. Visualization of this representation can be seen in ~\Figref{fig:data}, where the first three channels of audio data ($64\times64\times3$) are converted to an RGB image.

\begin{figure*}[t!]
\centering
\includegraphics[width=\textwidth]{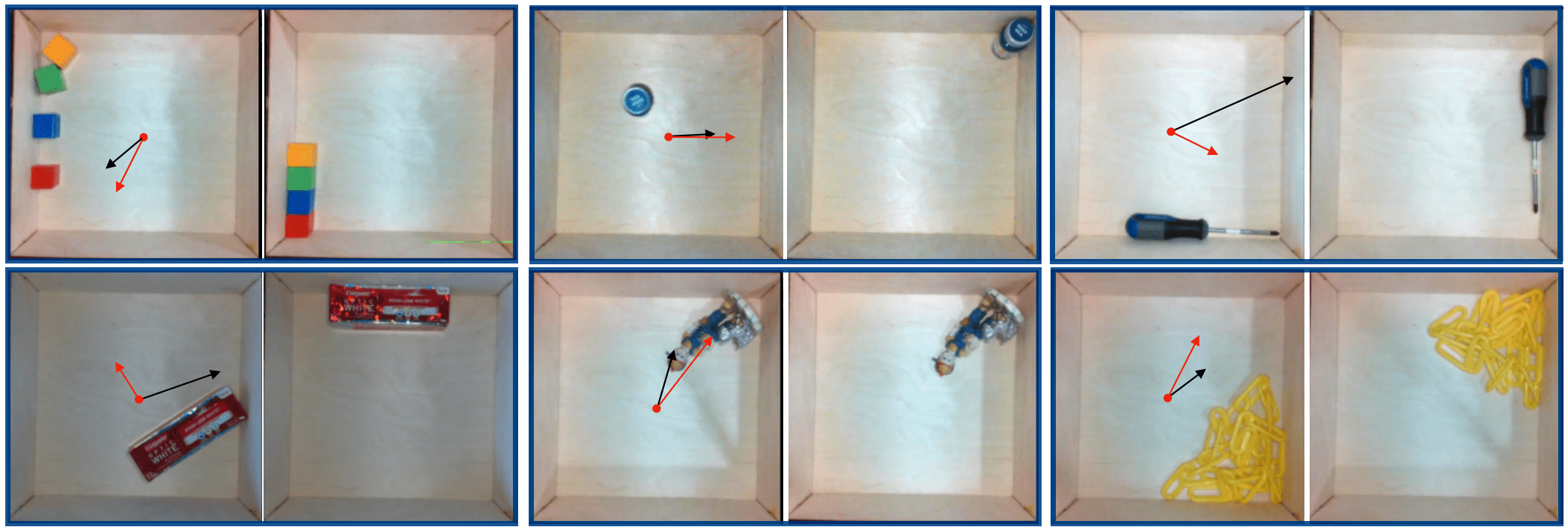}
\caption{Visualization of predictions from our inverse model. For each image, the left image is the start state, while the right image is the end state. The red arrow represents the ground truth actions taken by the robot, while the black arrow corresponds to the actions predicted by our action-regression model.}
\label{fig:inv_res}
\end{figure*}

\subsection{Fine-grained object classification}
\label{sec:class}
Classically, the goal of recognition is to identify which object is being perceived. This task is generally done using visual images as input, and is done to test the robustness of visual feature extractors. In our case, we use this task to study what type of object-centric information is contained in sound. For the 60 objects in our TiltBot dataset we first create a training set with 80\% of the data and a testing set with the remaining 20\%. Then, we train a simple CNN~\cite{krizhevsky2012imagenet}, that only takes the audio information as input and outputs the instance label of the object that generated the sound. This architecture is similar to top part of ~\Figref{fig:nets}(a). 

\begin{figure*}[t!]
\centering
\includegraphics[width=\textwidth]{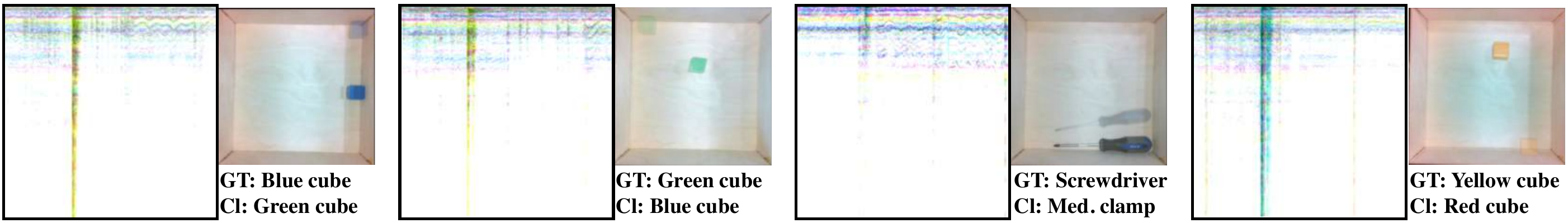}
\caption{Here we see the top classification errors made by our instance recognition model. There are two key failure modes: (a) misclassifications between similar objects that have different visual properties like the cubes, and (b) when the sound generated is too soft like the misclassification of the screwdriver.}
\label{fig:cl_errors}
\end{figure*}

On our heldout testing set, this trained model achieves a classification accuracy of 76.1\%. Note that a random classifier gets a 1.67\% accuracy. This shows that audio data contains fine-grained information about objects. 
Although \citet{owens2016visually} demonstrates that audio information can be used to classify broad categories like wood, metal etc., our results show for the first time (to our knowledge) that audio information generated through action gives instance-level information like screwdriver, scissor, tennis ball etc. 
To further understand what information sound gives us, we study the top classification errors of our model. In ~\Figref{fig:cl_errors} we see that there are two main modes of errors. The first is if instances only differ visually. For example, a green cube cannot be distinguished from a blue cube solely from the sound information. The second error mode is the generated sound is too soft. If the action causes the object to only move a little and not make too much sound, information about the object is masked away and causes classification errors.



\begin{table*}[t!]
\normalsize
\centering
\begin{tabular}{r|lllll|cccc|c|}
\multicolumn{1}{l|}{} & \multicolumn{5}{c|}{Train objects} & \multicolumn{5}{c|}{Test objects} \\ \hline
$\lambda=$ & \multicolumn{1}{c}{0.0} & \multicolumn{1}{c}{0.05} & \multicolumn{1}{c}{0.1} & \multicolumn{1}{c}{0.2} & \multicolumn{1}{c|}{1.0} & 0.0 & 0.1 & 0.2 & 1.0 & Image \\ \hline
Class. ($\uparrow$) & 0.738 & 0.780 & 0.770 & \textbf{0.786} & 0.027 & N/A & N/A & N/A & N/A & N/A \\
Reg. ($\downarrow$) & 0.395 & 0.024 & 0.022 & 0.014 & \textbf{0.008} & \multicolumn{1}{l}{0.352} & \multicolumn{1}{l}{0.027} & \multicolumn{1}{l}{\textbf{0.020}} & \multicolumn{1}{l|}{0.027} & \multicolumn{1}{l|}{0.043}
\end{tabular}
\caption{Classification and Regression performance across different methods on the Tilt-Bot dataset. For classification, higher is better while for regression lower is better.}
\label{tab:cl_reg_comp}
\end{table*}


\subsection{Inverse-model learning}
The goal of learning inverse models is to identify what action was applied, given observations before and after the action. From a biological perspective, learning inverse models implies an understanding of cause and effect and is often necessary for efficient motor-learning~\cite{wolpert1998multiple}. In the general setting of this problem, a model takes as input the observations before and after an interaction and outputs the action applied during the interaction. In our case, we want to study if sound contains cause-effect information about actions. Moreover, since inverse-model learning can be evaluated on previously unseen objects, we can test the generalization of audio features not only on objects seen in training but to novel objects as well. 

To demonstrate this, we split our TiltBot objects into two sets: set A and set B, where both sets contain 30 objects with 15 objects from the YCB dataset. Using an architecture similar to the bottom part of ~\Figref{fig:nets}(a), an inverse model is trained on set A to regress the action. The input into this inverse model is an image of the object before the interaction, and the sound generated during the interaction. Note that the image of the object after the interaction is not given as input. The action that needs to be output is the 2D projection of the rotation vector on the planar tray surface. We evaluate the performance of the inverse model using normalized ($\lbrack-1,1 \rbrack$) mean squared error (MSE), where lower is better. Testing this model on held-out set A objects, we get a MSE of 0.008, while a random model gives a MSE of 0.4. If we use the image of the object after the interaction as input instead of audio, we get a MSE of 0.043. This shows that for these physical interactions using audio information is not just better than random, but in fact better than using visual observations. This insight holds true even when tested on previously unseen set B objects. With set B testing, audio inverse models give a MSE of 0.027, which indicates some amount of overfitting on set A objects. However, this is significantly better than the 0.047 MSE we get from using purely visual inverse models. Sample evaluations of our inverse model can be seen in ~\Figref{fig:inv_res}.



\subsection{Multi-task audio embedding learning}
In the previous two sections, we have seen how sound contains information about both fine-grained instances of objects and causal effects of the action. But what is the right loss function to train an audio embedding that generalizes to multiple downstream tasks? One way would be to train the embedding on the instance recognition task on Tilt-Bot data, while another option would be to train it on the inverse-model task. Both of these tasks encode different forms of information, with classification encoding identifiable properties of the object and inverse model encoding the physical properties of the object. Inspired from work in multi-task learning~\cite{caruana1997multitask,pinto2016mlt}, we take the best of both worlds and train a joint embedding that can simultaneously encode both classification and action information.

As seen in ~\Figref{fig:nets}(a), the audio embedding $e_A$ is trained jointly using the classification and the inverse model loss according to $\mathcal{L}_{total} = (1-\lambda)\mathcal{L}_{class} + \lambda\mathcal{L}_{inv}$. Note that when $\lambda=0$, the embedding captures only classification information, while $\lambda=1$ captures only inverse-model information. We report the performance of joint learning on held-out data in \Tabref{tab:cl_reg_comp}. Here, training is performed on set A objects, while testing is done on set A held-out interactions and unseen set B objects. For classification, we find that joint learning improves performance from 73.8\% on the 30 set A objects to 78.6\%. When trained on both set A and set B objects, classification performance improves from 76.1\%~(\Secref{sec:class}) to 79.5\%. On inverse-model learning, we notice that joint learning does not improve performance on set A. However, on novel set B objects, we see a significant improvement from $0.027$ MSE to $0.020$ MSE. Again, this performance is also much better than learning directly from visual inverse-models at $0.043$ MSE.

\begin{figure}
\centering
\begin{minipage}{.5\linewidth}
  \begin{center}
        \includegraphics[width=\linewidth]{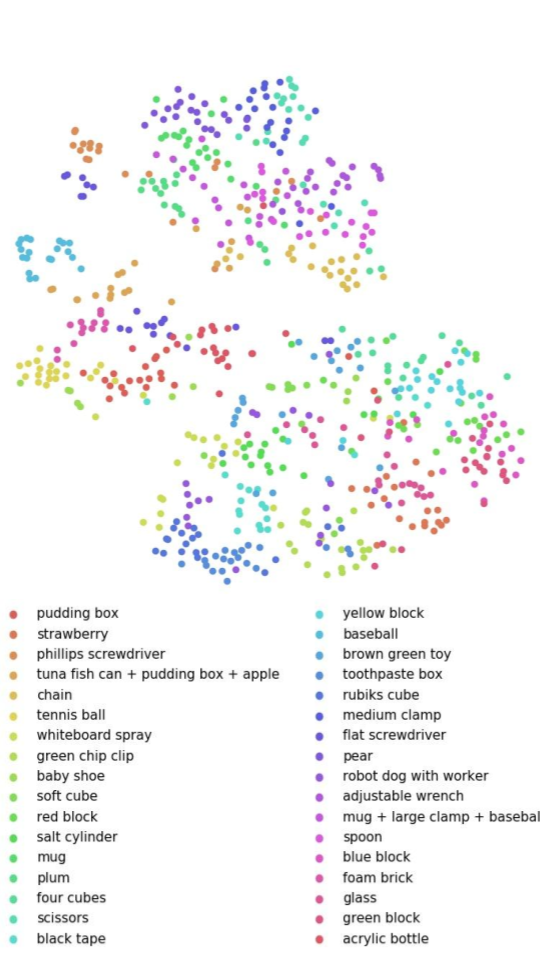}
        \caption*{(a)}
  \end{center}
\end{minipage}%
\begin{minipage}{.5\linewidth}
  \begin{center}
        \includegraphics[width=\linewidth]{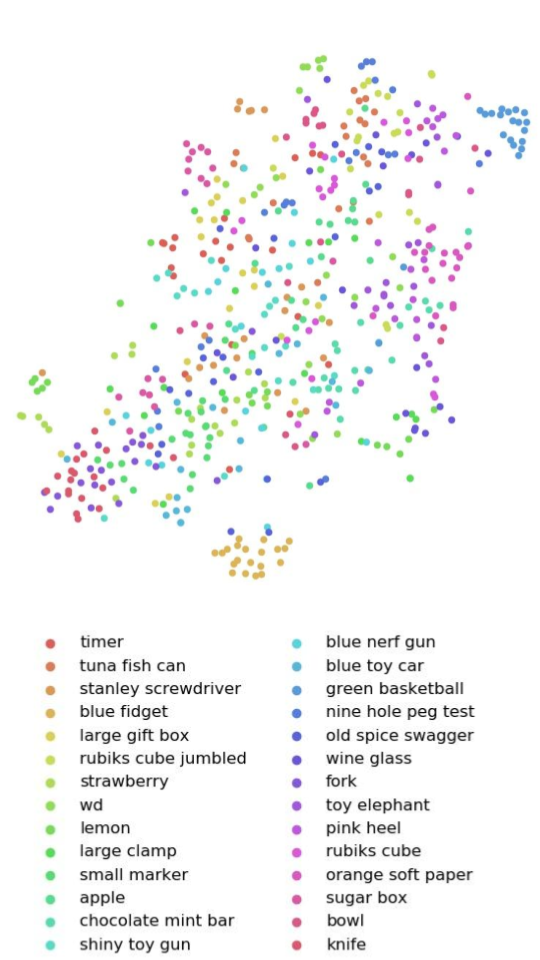}
        \caption*{(b)}
  \end{center}
\end{minipage}
\caption{t-SNE plots using audio encoding extracted from the audio embedding network. (a) represents the training objects and (b) for testing objects. This plot is best viewed in color.}
\label{fig:tsne}
\end{figure}



\begin{figure*}[t!]
\centering
\includegraphics[width=\linewidth]{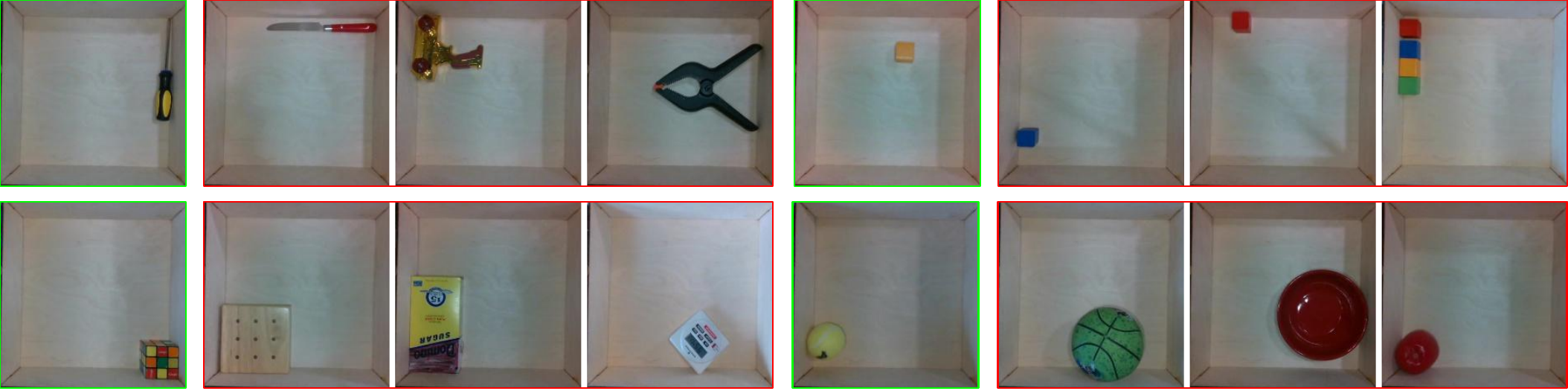}
\caption{Image retrieval results based on audio embedding. Here, the green box image corresponds to the query image and 3 images in the row (in red) correspond to the closest object retrievals in the audio-embedding space.}
\label{fig:img_retrieval}
\end{figure*}

Another way to understand the information captured in our audio embeddings is to look at the top three nearest object instance given an input object instance. In \Figref{fig:img_retrieval} we show a few of these object retrievals. Interestingly, these features capture object shapes like matching the long screwdriver to the long butterknife and matching the yellow cube to other colored cubes. In \Figref{fig:tsne} we show the tSNE~\cite{maaten2008visualizing} plots of the features. This further demonstrates how similar objects are closer, while physically different objects are farther apart.




\subsection{Few shot learning}
\begin{figure*}[t!]
\centering
\includegraphics[width=\linewidth]{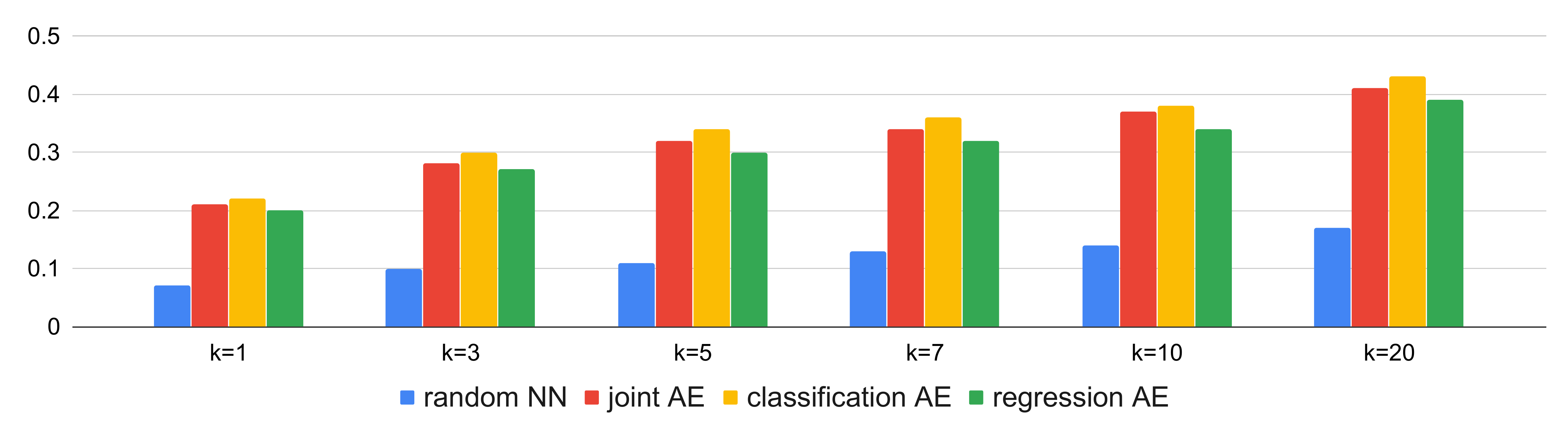}
\caption{Classification accuracy on test objects based on k nearest neighbors of various audio embeddings (AE). Here, we note that embeddings learned through our audio networks are significantly better than random embeddings.}
\label{fig:few_shot}
\end{figure*}

To further quantify the generalization of the trained audio embeddings, we perform a few shot learning experiment~\cite{fewshot2006}. We first extract the embeddings corresponding to novel objects in set B using the embedding model learned on set A. Given a few $k$ interactions of an object, we classify the object belonging to test B set using nearest neighbours on the audio features. Results for this are visualized in ~\Figref{fig:few_shot}. All audio embeddings achieve similar performance on this task going from around $25\%$ with $k=1$ to around $40\%$ with $k=20$. We note that although the performance is more than 2X of using a random neural-network audio embedding, it significantly worse than using ImageNet pretrained ResNet~\cite{he2016deep} features that have an impressive 75\% for $k=1$. This shows that for pure visual similarity tasks, audio information is helpful, but currently not informative enough compared to state-of-the-art visual features that are trained on a significantly larger amount of visual data.

\subsection{Downstream Task: Forward model learning on TiltBot}
Our previous experiments demonstrate the importance of using audio perception. In this section, we investigate if we can use sound to extract physical properties of an object before physically interacting with it. This use case is inspired from recent work on Environment Probing Interactions~\cite{zhou2019environment} where probing interactions are used to understand latent factors before implementing the real policy. Here the sound generated through probing interactions would serve as latent parameters for representing the object. 

To evaluate the use of audio features for downstream tasks, we perform forward prediction (See ~\Figref{fig:nets}(b)). Here given an object, a random interaction is performed on it and a sound is generated from this interaction. The embedding network trained using multi-task learning is then used to extract the audio embedding, which will serve as our object's representation. Then, given this object's representation, we can train a forward model that takes as additional input the location of the object and action applied on the object and outputs the location of the object after the interaction. To learn this forward model, the network has to understand the dynamics of the object. Note that the only object specific information is given through the audio embedding. 


As seen in ~\Tabref{tab:comp_fm}, we report significant improvements in forward model prediction from 0.258 L1 error using visual features to 0.220 L1 error when using the audio embedding feature on objects seen during forward model training. This trend continues for novel set B objects, where both the embedding and forward model was trained on set A objects. Here we see an even larger improvement with visual features giving 0.256 L1 error while audio features giving 0.193 L1 error. This shows that audio embedding information is better able to capture implicit physics of the object as compared to visual features. Moreover, these features are significantly more robust than visual features and also generalize to previously unseen objects.

\begin{figure}[t!]
\centering
\includegraphics[width=\linewidth]{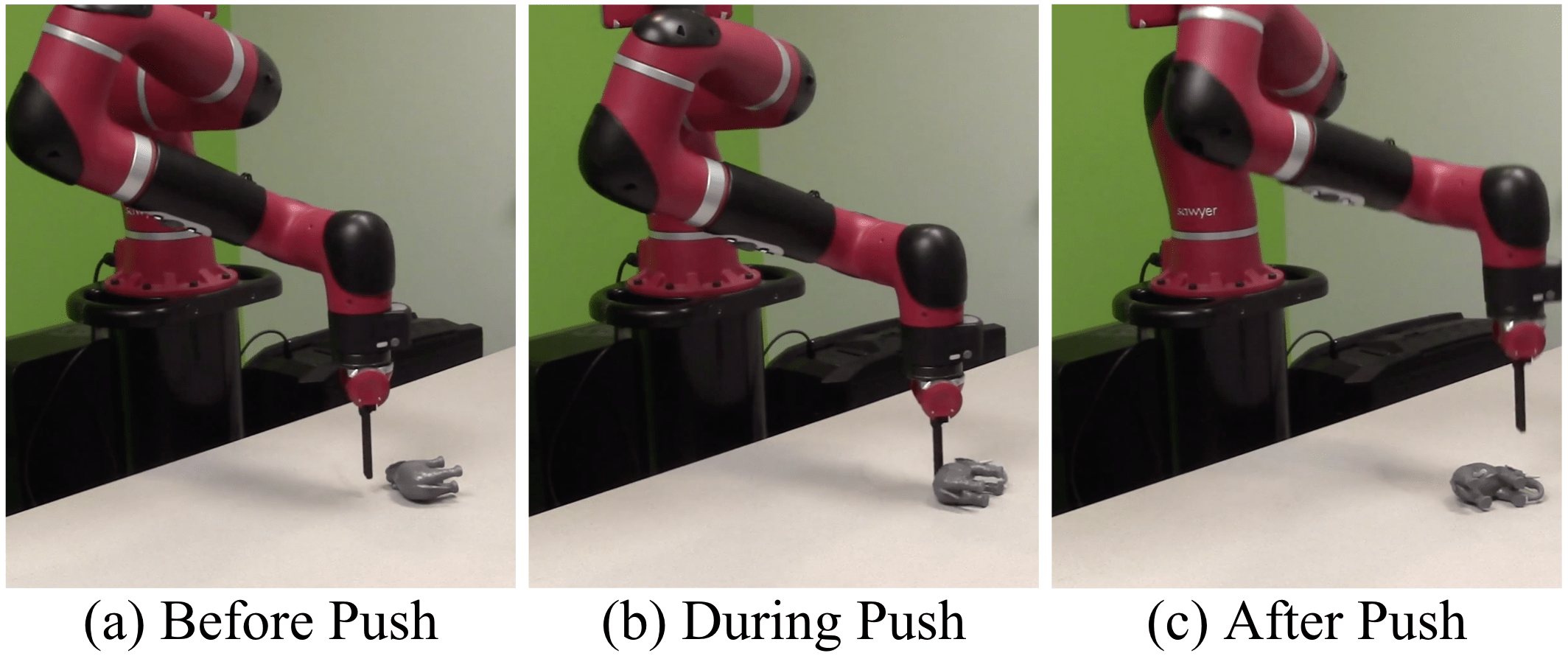}
\caption{Setup for collecting table-top pushing data. (a) represents the state of the robot before pushing an object, (b) while pushing an object, and (c) after pushing an object. We record the image of the table before and after pushing the object along with the action taken by the robot.}
\label{fig:pushing}
\end{figure}

\begin{table*}[t!]
\normalsize
\centering
\begin{tabular}{r|ccccc|ccc}
\multicolumn{1}{l|}{} & \multicolumn{5}{c|}{Audio Embeddings} & \multicolumn{3}{c}{No Audio} \\ \hline
$\lambda =$ & 0 & 0.05 & 0.1 & 0.2 & 1.0 & ResNet & Oracle \\ \hline
Train Objects & 0.225 & 0.221 & \textbf{0.220} & 0.222 & 0.239 & 0.258  & 0.206 \\
Test Objects & 0.195 & 0.194 & \textbf{0.193} & 0.1945 & 0.195 & 0.256  & 0.155
\end{tabular}
\caption{Comparison of using audio embeddings versus pre-trained visual embeddings for forward model prediction. Oracle represents training with object class labels as input.}
\label{tab:comp_fm}
\end{table*}

\begin{figure*}[t!]
\centering
\includegraphics[width=\linewidth]{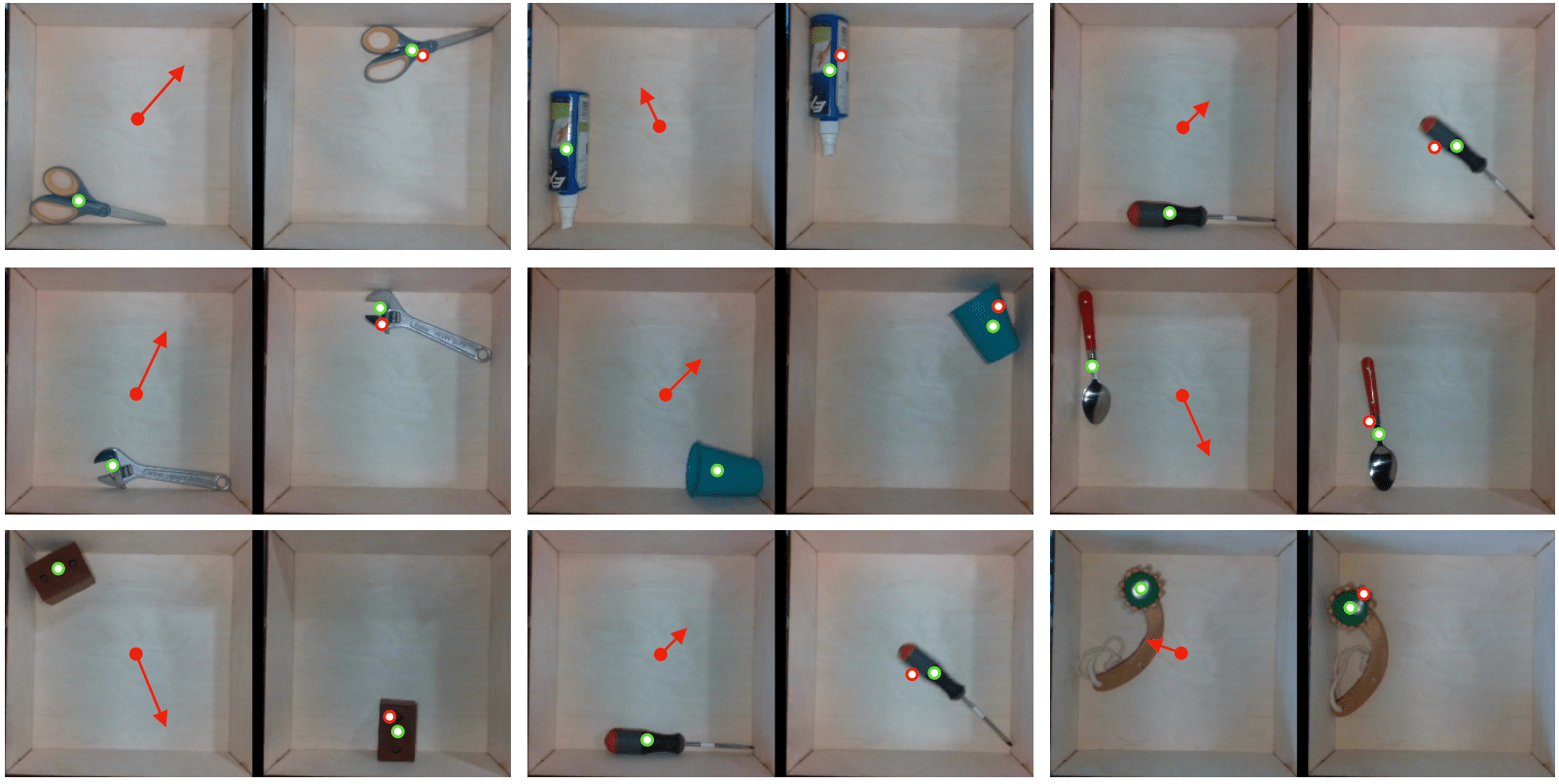}
\caption{Forward model predictions are visualized here as pairs of images. The left image is the observation before the interaction, while the right image is the observation after the interaction. Based on the object ground truth location (shown as the green dot) before interaction, the audio embedding of the object and action taken by the robot (shown as a red arrow), trained forward model predicts the future object location (shown as a red dot).}
\label{fig:forward_model}
\end{figure*}

\subsection{Downstream Task: Forward model learning on robotic pushing} 
To evaluate how audio embeddings can be useful to perform tasks outside the tray environment, we perform a table top pushing experiment. For this, we use a Sawyer robot to collect push data. Given an object placed on the robot’s table, we first perform background subtraction to determine the position of the object. Then, based on this position we sample two points such that line joining them will pass through the object. We record images $I_{start}$ and $I_{end}$ to capture the state of the table before and after push action being applied by the robot. In addition to this, we also record position $P_{start}$ and $P_{end}$ which depicts the starting and ending push location of the robot end-effector. The robot's motion $\{P_{start}, P_{end}\}$ serves as the pushing action. Visualization of this setup can be seen in~\Figref{fig:pushing}. Using this setup we collect a dataset of 1000 planar pushing interactions on 10 training set and 10 testing set objects. Given this data, we learn an audio embedding conditioned forward model similar to \Figref{fig:nets}(b). The model predicts the location of the object in $I_{end}$ given object location in $I_{start}$, the robots action $\{P_{start}, P_{end}\}$, and an audio embedding of the object from a prior TiltBot interaction. We note that without audio embedding the L2 error of the pushing location on test-set objects is 0.180 (normalized coordinates) while using audio embeddings gives a reduced L2 error of 0.159. Although performed in the relaxed state-based setting, this experiment demonstrates the promise of audio embeddings in improving robotic tasks like pushing. 


\section{Conclusion}
In this work, we perform one of the first studies on the interactions between sound and action. Through our sound-action-vision dataset collected using our Tilt-Bot robot, we present several insights into what information can be extracted from sound. From fine-grained object recognition to inverse-model learning, we demonstrate that sound can provide valuable information that can be used in downstream motor-control or robotic tasks. In some domains like forward model learning, we show that sound in fact provides more information than can be obtained from visual information alone. We hope that the Tilt-Bot dataset, which will be publicly released, along with our findings will inspire future work in the sound-action domain and find widespread applicability in robotics.
\vspace{0.1in}

\noindent \textbf{Acknowledgments:} We thank DARPA MCS, ONR MURI and an ONR Young Investigator award for funding this work. We also thank Xiaolong Wang and Olivia Watkins for insightful comments and discussions.

\bibliographystyle{plainnat}
\bibliography{references}

\begin{thebibliography}{45}
\providecommand{\natexlab}[1]{#1}
\providecommand{\url}[1]{\texttt{#1}}
\expandafter\ifx\csname urlstyle\endcsname\relax
  \providecommand{\doi}[1]{doi: #1}\else
  \providecommand{\doi}{doi: \begingroup \urlstyle{rm}\Url}\fi

\bibitem[cam()]{camera}
Intel realsense d435 rgbd camera.
\newblock
  \url{https://click.intel.com/intelr-realsensetm-depth-camera-d435.html}.

\bibitem[mic()]{microphone}
Contact microphone.
\newblock
  \url{https://www.amazon.com/Neewer-Contact-Microphone-Mandolin-Accurate/dp/B019TW4BZO}.

\bibitem[Agrawal et~al.(2016)Agrawal, Nair, Abbeel, Malik, and
  Levine]{agarwal2016}
Pulkit Agrawal, Ashvin Nair, Pieter Abbeel, Jitendra Malik, and Sergey Levine.
\newblock Learning to poke by poking: Experiential learning of intuitive
  physics.
\newblock \emph{NIPS}, 2016.

\bibitem[Amos et~al.(2018)Amos, Dinh, Cabi, Roth{\"o}rl, Colmenarejo, Muldal,
  Erez, Tassa, de~Freitas, and Denil]{amos2018learning}
Brandon Amos, Laurent Dinh, Serkan Cabi, Thomas Roth{\"o}rl, Sergio~G{\'o}mez
  Colmenarejo, Alistair Muldal, Tom Erez, Yuval Tassa, Nando de~Freitas, and
  Misha Denil.
\newblock Learning awareness models.
\newblock \emph{arXiv preprint arXiv:1804.06318}, 2018.

\bibitem[Arandjelovic and
  Zisserman(2017{\natexlab{a}})]{andrew_look_listen_learn}
Relja Arandjelovic and Andrew Zisserman.
\newblock Look, listen and learn.
\newblock \emph{CoRR}, abs/1705.08168, 2017{\natexlab{a}}.
\newblock URL \url{http://arxiv.org/abs/1705.08168}.

\bibitem[Arandjelovic and
  Zisserman(2017{\natexlab{b}})]{andrew_objects_that_sound}
Relja Arandjelovic and Andrew Zisserman.
\newblock Objects that sound.
\newblock \emph{CoRR}, abs/1712.06651, 2017{\natexlab{b}}.
\newblock URL \url{http://arxiv.org/abs/1712.06651}.

\bibitem[Aytar et~al.(2016)Aytar, Vondrick, and Torralba]{carl_soundnet}
Yusuf Aytar, Carl Vondrick, and Antonio Torralba.
\newblock Soundnet: Learning sound representations from unlabeled video.
\newblock In D.~D. Lee, M.~Sugiyama, U.~V. Luxburg, I.~Guyon, and R.~Garnett,
  editors, \emph{Advances in Neural Information Processing Systems 29}, pages
  892--900. Curran Associates, Inc., 2016.

\bibitem[Calli et~al.(2015)Calli, Walsman, Singh, Srinivasa, Abbeel, and
  Dollar]{calli2015benchmarking}
Berk Calli, Aaron Walsman, Arjun Singh, Siddhartha Srinivasa, Pieter Abbeel,
  and Aaron~M Dollar.
\newblock Benchmarking in manipulation research: The ycb object and model set
  and benchmarking protocols.
\newblock \emph{arXiv preprint arXiv:1502.03143}, 2015.

\bibitem[Caruana(1997)]{caruana1997multitask}
Rich Caruana.
\newblock Multitask learning.
\newblock \emph{Machine learning}, 28\penalty0 (1):\penalty0 41--75, 1997.

\bibitem[Clarke et~al.(2018)Clarke, Rhodes, Atkeson, and
  Kroemer]{pmlr-v87-clarke18a}
Samuel Clarke, Travers Rhodes, Christopher~G. Atkeson, and Oliver Kroemer.
\newblock Learning audio feedback for estimating amount and flow of granular
  material.
\newblock In Aude Billard, Anca Dragan, Jan Peters, and Jun Morimoto, editors,
  \emph{Proceedings of The 2nd Conference on Robot Learning}, volume~87 of
  \emph{Proceedings of Machine Learning Research}, pages 529--550. PMLR, 29--31
  Oct 2018.
\newblock URL \url{http://proceedings.mlr.press/v87/clarke18a.html}.

\bibitem[Cosgun et~al.(2011)Cosgun, Hermans, Emeli, and
  Stilman]{cosgun2011push}
Akansel Cosgun, Tucker Hermans, Victor Emeli, and Mike Stilman.
\newblock Push planning for object placement on cluttered table surfaces.
\newblock In \emph{2011 IEEE/RSJ international conference on intelligent robots
  and systems}, pages 4627--4632. IEEE, 2011.

\bibitem[Daubechies(1990)]{daubechies1990wavelet}
Ingrid Daubechies.
\newblock The wavelet transform, time-frequency localization and signal
  analysis.
\newblock \emph{IEEE transactions on information theory}, 36\penalty0
  (5):\penalty0 961--1005, 1990.

\bibitem[Deisenroth and Rasmussen(2011)]{deisenroth2011pilco}
Marc Deisenroth and Carl~E Rasmussen.
\newblock Pilco: A model-based and data-efficient approach to policy search.
\newblock In \emph{Proceedings of the 28th International Conference on machine
  learning (ICML-11)}, pages 465--472, 2011.

\bibitem[Dogar and Srinivasa(2012)]{dogar2012planning}
Mehmet~R Dogar and Siddhartha~S Srinivasa.
\newblock A planning framework for non-prehensile manipulation under clutter
  and uncertainty.
\newblock \emph{Autonomous Robots}, 33\penalty0 (3):\penalty0 217--236, 2012.

\bibitem[Ebert et~al.(2018)Ebert, Finn, Dasari, Xie, Lee, and
  Levine]{ebert2018visual}
Frederik Ebert, Chelsea Finn, Sudeep Dasari, Annie Xie, Alex Lee, and Sergey
  Levine.
\newblock Visual foresight: Model-based deep reinforcement learning for
  vision-based robotic control.
\newblock \emph{arXiv preprint arXiv:1812.00568}, 2018.

\bibitem[Fei-Fei et~al.(2006)Fei-Fei, Fergus, and Perona]{fewshot2006}
Li~Fei-Fei, Rob Fergus, and Pietro Perona.
\newblock One-shot learning of object categories.
\newblock \emph{IEEE Trans. Pattern Anal. Mach. Intell.}, 28\penalty0
  (4):\penalty0 594--611, April 2006.
\newblock ISSN 0162-8828.
\newblock \doi{10.1109/TPAMI.2006.79}.
\newblock URL \url{https://doi.org/10.1109/TPAMI.2006.79}.

\bibitem[Finn and Levine(2017)]{finn2017deep}
Chelsea Finn and Sergey Levine.
\newblock Deep visual foresight for planning robot motion.
\newblock In \emph{2017 IEEE International Conference on Robotics and
  Automation (ICRA)}, pages 2786--2793. IEEE, 2017.

\bibitem[Gal et~al.(2016)Gal, McAllister, and Rasmussen]{gal2016improving}
Yarin Gal, Rowan McAllister, and Carl~Edward Rasmussen.
\newblock Improving {PILCO} with {B}ayesian neural network dynamics models.
\newblock In \emph{Data-Efficient Machine Learning workshop, International
  Conference on Machine Learning}, 2016.

\bibitem[Gandhi et~al.(2017)Gandhi, Pinto, and
  Gupta]{DBLP:journals/corr/GandhiPG17}
Dhiraj Gandhi, Lerrel Pinto, and Abhinav Gupta.
\newblock Learning to fly by crashing.
\newblock \emph{CoRR}, abs/1704.05588, 2017.
\newblock URL \url{http://arxiv.org/abs/1704.05588}.

\bibitem[Hafner et~al.(2018)Hafner, Lillicrap, Fischer, Villegas, Ha, Lee, and
  Davidson]{hafner2018learning}
Danijar Hafner, Timothy Lillicrap, Ian Fischer, Ruben Villegas, David Ha,
  Honglak Lee, and James Davidson.
\newblock Learning latent dynamics for planning from pixels.
\newblock \emph{arXiv preprint arXiv:1811.04551}, 2018.

\bibitem[He et~al.(2016)He, Zhang, Ren, and Sun]{he2016deep}
Kaiming He, Xiangyu Zhang, Shaoqing Ren, and Jian Sun.
\newblock Deep residual learning for image recognition.
\newblock In \emph{Proceedings of the IEEE conference on computer vision and
  pattern recognition}, pages 770--778, 2016.

\bibitem[He et~al.(2017)He, Gkioxari, Dollar, and Girshick]{He_2017_ICCV}
Kaiming He, Georgia Gkioxari, Piotr Dollar, and Ross Girshick.
\newblock Mask r-cnn.
\newblock In \emph{The IEEE International Conference on Computer Vision
  (ICCV)}, Oct 2017.

\bibitem[Henaff et~al.(2017)Henaff, Whitney, and LeCun]{henaff2017model}
Mikael Henaff, William~F Whitney, and Yann LeCun.
\newblock Model-based planning with discrete and continuous actions.
\newblock \emph{arXiv preprint arXiv:1705.07177}, 2017.

\bibitem[Kar et~al.(2015)Kar, Tulsiani, Carreira, and Malik]{shapesKarTCM15}
Abhishek Kar, Shubham Tulsiani, Jo{\~{a}}o Carreira, and Jitendra Malik.
\newblock Category-specific object reconstruction from a single image.
\newblock In \emph{Computer Vision and Pattern Regognition (CVPR)}, 2015.

\bibitem[Khatib(1987)]{Khatib1987AUA}
Oussama Khatib.
\newblock A unified approach for motion and force control of robot
  manipulators: The operational space formulation.
\newblock \emph{IEEE J. Robotics and Automation}, 3:\penalty0 43--53, 1987.

\bibitem[Krizhevsky et~al.(2012)Krizhevsky, Sutskever, and
  Hinton]{krizhevsky2012imagenet}
Alex Krizhevsky, Ilya Sutskever, and Geoffrey~E Hinton.
\newblock Imagenet classification with deep convolutional neural networks.
\newblock In \emph{NIPS}, 2012.

\bibitem[Levine et~al.(2016{\natexlab{a}})Levine, Finn, Darrell, and
  Abbeel]{levine2016end}
Sergey Levine, Chelsea Finn, Trevor Darrell, and Pieter Abbeel.
\newblock End-to-end training of deep visuomotor policies.
\newblock \emph{JMLR}, 2016{\natexlab{a}}.

\bibitem[Levine et~al.(2016{\natexlab{b}})Levine, Pastor, Krizhevsky, and
  Quillen]{levine2016learning}
Sergey Levine, Peter Pastor, Alex Krizhevsky, and Deirdre Quillen.
\newblock Learning hand-eye coordination for robotic grasping with deep
  learning and large-scale data collection.
\newblock \emph{ISER}, 2016{\natexlab{b}}.

\bibitem[Maaten and Hinton(2008)]{maaten2008visualizing}
Laurens van~der Maaten and Geoffrey Hinton.
\newblock Visualizing data using t-sne.
\newblock \emph{Journal of machine learning research}, 9\penalty0
  (Nov):\penalty0 2579--2605, 2008.

\bibitem[Murali et~al.(2018)Murali, Li, Gandhi, and
  Gupta]{DBLP:journals/corr/abs-1805-04201}
Adithyavairavan Murali, Yin Li, Dhiraj Gandhi, and Abhinav Gupta.
\newblock Learning to grasp without seeing.
\newblock \emph{CoRR}, abs/1805.04201, 2018.
\newblock URL \url{http://arxiv.org/abs/1805.04201}.

\bibitem[Murray(2017)]{murray2017mathematical}
Richard~M Murray.
\newblock \emph{A mathematical introduction to robotic manipulation}.
\newblock CRC press, 2017.

\bibitem[Owens et~al.(2015)Owens, Isola, McDermott, Torralba, Adelson, and
  Freeman]{andrew_sound}
Andrew Owens, Phillip Isola, Josh~H. McDermott, Antonio Torralba, Edward~H.
  Adelson, and William~T. Freeman.
\newblock Visually indicated sounds.
\newblock \emph{CoRR}, abs/1512.08512, 2015.
\newblock URL \url{http://arxiv.org/abs/1512.08512}.

\bibitem[Owens et~al.(2016)Owens, Isola, McDermott, Torralba, Adelson, and
  Freeman]{owens2016visually}
Andrew Owens, Phillip Isola, Josh McDermott, Antonio Torralba, Edward~H
  Adelson, and William~T Freeman.
\newblock Visually indicated sounds.
\newblock In \emph{Proceedings of the IEEE conference on computer vision and
  pattern recognition}, pages 2405--2413, 2016.

\bibitem[Pinto and Gupta(2016{\natexlab{a}})]{pinto2016mlt}
Lerrel Pinto and Abhinav Gupta.
\newblock Learning to push by grasping: Using multiple tasks for effective
  learning.
\newblock \emph{arXiv preprint arXiv:1609.09025}, 2016{\natexlab{a}}.

\bibitem[Pinto and Gupta(2016{\natexlab{b}})]{pinto2016supersizing}
Lerrel Pinto and Abhinav Gupta.
\newblock Supersizing self-supervision: Learning to grasp from 50k tries and
  700 robot hours.
\newblock \emph{ICRA}, 2016{\natexlab{b}}.

\bibitem[Pinto et~al.(2016)Pinto, Gandhi, Han, Park, and
  Gupta]{pinto2016curious}
Lerrel Pinto, Dhiraj Gandhi, Yuanfeng Han, Yong-Lae Park, and Abhinav Gupta.
\newblock The curious robot: Learning visual representations via physical
  interactions.
\newblock \emph{ECCV}, 2016.

\bibitem[Schneider et~al.(2009)Schneider, Sturm, Stachniss, Reisert, Burkhardt,
  and Burgard]{schneider2009object}
Alexander Schneider, J{\"u}rgen Sturm, Cyrill Stachniss, Marco Reisert, Hans
  Burkhardt, and Wolfram Burgard.
\newblock Object identification with tactile sensors using bag-of-features.
\newblock In \emph{IROS}, volume~9, pages 243--248, 2009.

\bibitem[Senocak et~al.(2018)Senocak, Oh, Kim, Yang, and
  Kweon]{DBLP:journals/corr/abs-1803-03849}
Arda Senocak, Tae{-}Hyun Oh, Jun{-}Sik Kim, Ming{-}Hsuan Yang, and In~So Kweon.
\newblock Learning to localize sound source in visual scenes.
\newblock \emph{CoRR}, abs/1803.03849, 2018.
\newblock URL \url{http://arxiv.org/abs/1803.03849}.

\bibitem[Wolpert and Kawato(1998)]{wolpert1998multiple}
Daniel~M Wolpert and Mitsuo Kawato.
\newblock Multiple paired forward and inverse models for motor control.
\newblock \emph{Neural networks}, 11\penalty0 (7-8):\penalty0 1317--1329, 1998.

\bibitem[Xiang et~al.(2015)Xiang, Alahi, and Savarese]{xiang2015learning}
Yu~Xiang, Alexandre Alahi, and Silvio Savarese.
\newblock Learning to track: Online multi-object tracking by decision making.
\newblock In \emph{International Conference on Computer Vision (ICCV)}, pages
  4705--4713, 2015.

\bibitem[Yu et~al.(2016)Yu, Bauza, Fazeli, and Rodriguez]{yu2016more}
Kuan-Ting Yu, Maria Bauza, Nima Fazeli, and Alberto Rodriguez.
\newblock More than a million ways to be pushed. a high-fidelity experimental
  dataset of planar pushing.
\newblock In \emph{2016 IEEE/RSJ international conference on intelligent robots
  and systems (IROS)}, pages 30--37. IEEE, 2016.

\bibitem[Zhang et~al.(2017)Zhang, Li, Huang, Wu, Tenenbaum, and
  Freeman]{zhang2017shape}
Zhoutong Zhang, Qiujia Li, Zhengjia Huang, Jiajun Wu, Josh Tenenbaum, and Bill
  Freeman.
\newblock Shape and material from sound.
\newblock In \emph{Advances in Neural Information Processing Systems}, pages
  1278--1288, 2017.

\bibitem[Zhao et~al.(2018)Zhao, Gan, Rouditchenko, Vondrick, McDermott, and
  Torralba]{Zhao_2018_ECCV}
Hang Zhao, Chuang Gan, Andrew Rouditchenko, Carl Vondrick, Josh McDermott, and
  Antonio Torralba.
\newblock The sound of pixels.
\newblock In \emph{The European Conference on Computer Vision (ECCV)},
  September 2018.

\bibitem[Zhou et~al.(2019)Zhou, Pinto, and Gupta]{zhou2019environment}
Wenxuan Zhou, Lerrel Pinto, and Abhinav Gupta.
\newblock Environment probing interaction policies.
\newblock \emph{ICLR}, 2019.

\bibitem[Zivkovic(2004)]{zivkovic2004improved}
Zoran Zivkovic.
\newblock Improved adaptive gaussian mixture model for background subtraction.
\newblock In \emph{Proceedings of the 17th International Conference on Pattern
  Recognition, 2004. ICPR 2004.}, volume~2, pages 28--31. IEEE, 2004.

\end{thebibliography}

\end{document}